\documentclass[nohyperref]{article}

\usepackage{amsmath,amsfonts,bm}

\def\eqref#1{equation~\ref{#1}}
\def\1{\bm{1}}

\DeclareMathAlphabet{\mathsfit}{\encodingdefault}{\sfdefault}{m}{sl}
\SetMathAlphabet{\mathsfit}{bold}{\encodingdefault}{\sfdefault}{bx}{n}

\newtheorem{lem}{Lemma}
\usepackage{listings}

\usepackage{algorithm}
\usepackage{algpseudocode}
\usepackage{microtype}
\usepackage{graphicx}
\usepackage{subfigure}
\usepackage{booktabs} % for professional tables

\usepackage{hyperref}

\newcommand{\soft}{SoftEdge}

\usepackage[accepted]{icml2022}
\usepackage{marginnote}

\icmltitlerunning{Regularizing  Graph Classification with Random Soft Edges}
\begin{document}

\twocolumn[
%\icmltitle{Towards  Intrusion-Free Graph Data Augmentation  for Graph Classification}
%\icmltitle{\soft: Combating Under-Fitting with  Connectivity-Preserving Graph Augmentation}
\icmltitle{\soft: Regularizing Graph Classification with Random Soft Edges}
%\icmltitle{Avoiding Manifold Intrusion in  Data Augmentation for Graph Classification}
%\icmltitle{Manifold Intrusion in  Graph Data Augmentation for Graph Classification}

%\icmltitle{Intrusion-Free Graph Augmentation for Graph Classification}
%\icmltitle{Avoiding Manifold Intrusion in Graph Data Augmentation}
%\icmltitle{Improving Graph Classification with Randomly Weighted Edges}



\begin{icmlauthorlist}
\icmlauthor{ Hongyu Guo}{yyy} 
\icmlauthor{Sun Sun}{yyy}
\end{icmlauthorlist}

\icmlaffiliation{yyy}{  National Research Council Canada,   1200 Montreal Road, Ottawa, email: firstname.lastname@nrc-cnrc.gc.ca}

%\icmlcorrespondingauthor{Hongyu Guo}{hongyu.guo@nrc-cnrc.gc.ca}
%\icmlcorrespondingauthor{Firstname2 Lastname2}{first2.last2@www.uk}

\icmlkeywords{Machine Learning, ICML}

\vskip 0.3in
]



\printAffiliationsAndNotice{}  % leave blank if no need to mention equal contribution

\begin{abstract}
Augmented graphs play a vital role in regularizing Graph Neural Networks (GNNs), which leverage information exchange along edges in  graphs,  in the form of message passing, for learning.   Due to their   effectiveness,  simple edge  and node manipulations (e.g., addition and deletion)  have been widely used in graph augmentation. Nevertheless, such  common augmentation techniques can  dramatically change the semantics of the original graph, causing overaggressive  augmentation and thus  under-fitting in the GNN learning. To address this problem 
\textcolor{black}{arising from dropping or adding graph edges and nodes}, we propose  \soft$ $, which  assigns random  weights to a portion of the edges of a given graph for augmentation. The synthetic graph generated by  \soft$ $ maintains the same nodes and their connectivities as the original graph, thus mitigating the semantic changes of the original graph.  We empirically   show that  this simple method obtains  superior  accuracy to popular  node and edge manipulation approaches and notable resilience to the accuracy degradation with the  GNN depth. 
\end{abstract}

%\begin{wrapfigure}{r}{0.4\textwidth}	
 \begin{figure}[h]
%\vspace{-5mm}
	\centering
	{\includegraphics[width=0.3\textwidth]{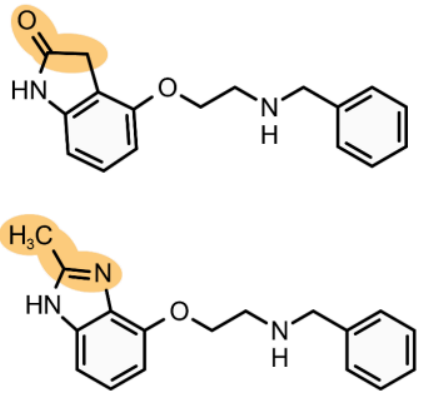}}	
%\vspace{-3mm}
	\caption{A pair of molecular graphs exhibit the activity cliff phenomena~\citep{Tilborg2022}: highly similar structure but  dramatic  differences in potency. }
%\vspace{-2mm}
	\label{fig:intrusion:nci1}
\end{figure}
%\end{wrapfigure}
\section{Introduction}
\label{intro}
Graph Neural Networks (GNNs)~\citep{kipf2017semi,velickovic2018graph}, which  iteratively propagate learned information  
in the form of message passing, 
have recently emerged as powerful approaches 
on a wide variety of tasks, including drug discovery~\citep{stokes2020deep}, 
  chip design~\citep{circuit-gnn}, and catalysts invention~\citep{godwin2021deep}. Recent studies on GNNs, nevertheless, also reveal a challenge in training GNNs. That is, similar to other successfully deployed deep neural networks, 
 GNNs  also require strong model regularization techniques to rectify their 
  over-parameterized learning paradigm.

 %To cope with the above challenge, 
 To this end, various regularization techniques for GNNs have been actively investigated, combating  issues such as over-fitting~\citep{10.1145/3446776}, over-smoothing~\citep{LiHW18,abs-1901-00596}, and  over-squashing~\citep{alon2021on}. 
% for better  model generalization. 
Despite the complexity of arbitrary structure and topology in graph data,  simple edge and node manipulations (e.g., addition and deletion) on  graphs~\citep{rong2020dropedge,3412086zhou,You2020GraphCL,zhao2021data,papp2021dropgnn} represent a very effective data augmentation strategy, which has been widely used to regularize the learning of GNNs.  

%\begin{wrapfigure}{r}{0.45\textwidth}
 \begin{figure}[h]
 %\vspace{-2mm}
	\centering
	{\includegraphics[width=0.45\textwidth]{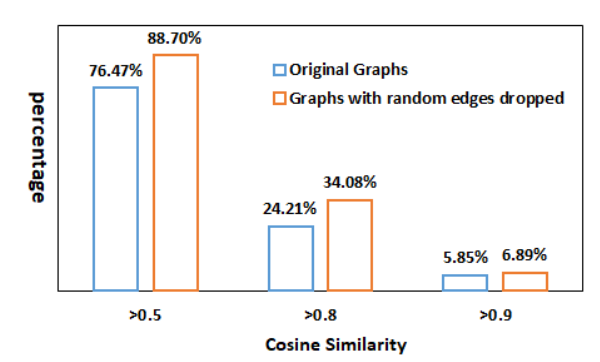}}	
%\vspace{-5mm}
	\caption{Percentages of graph pairs  with certain cosine similarity score but with opposite labels in the NCI1 dataset. This serves as a motivation for careful graph topology modification when conducting graph augmentation for model regularization. }
%\vspace{-2mm}
	\label{fig:similarity:nci1}
\end{figure}
%\end{wrapfigure}

In this paper, we reveal an overlooked issue in the aforementioned graph data augmentation technique for supervised graph classification. That is, simple edge and node manipulations such as addition and deletion can  
dramatically change the semantics of the original graphs. Consequently, regualarization with such overaggressive graph augmentation methods 
 essentially induces a form of under-fitting, which would inevitably result in the performance degradation of the model. 
For example,  activity cliff phenomena~\citep{articleMaggiora,Tilborg2022} has been well observed in molecular graphs. That is, a pair of molecules have highly similar structure but exhibit dramatic  differences in potency. 
Figure~\ref{fig:intrusion:nci1} visualizes a pair of such molecules~\citep{Tilborg2022}. 
For those molecular pairs,  edge or node operation on the graphs risks shifting the functionalities of the molecules, causing overaggressive  augmentation.

%\begin{wrapfigure}{r}{0.45\textwidth}
\begin{figure}%
%\vspace{-9mm}
	\centering
	{\includegraphics[width=0.45\textwidth]{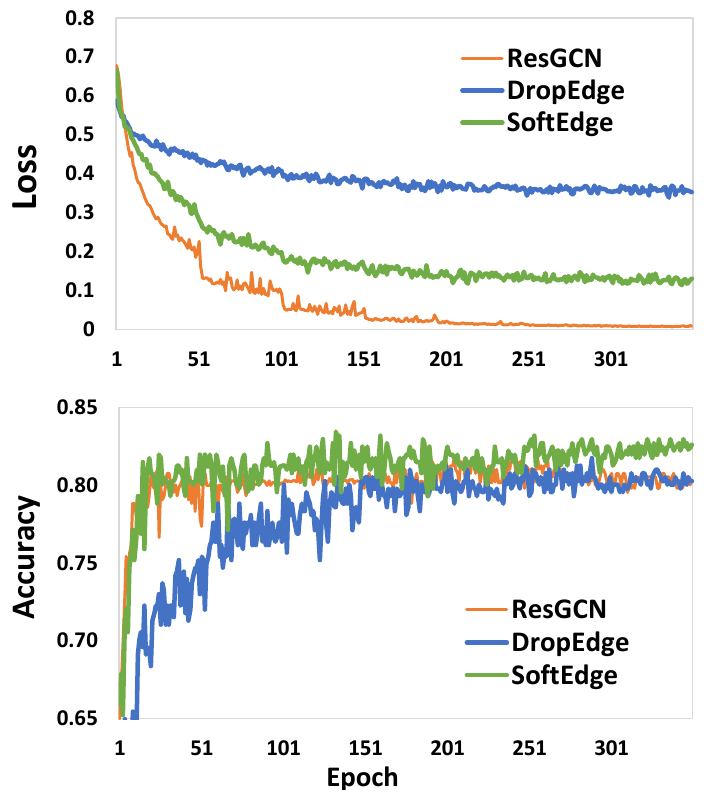}}	

%\vspace{-3mm}
	\caption{Training loss (top) and validation accuracy (bottom), obtained by ResGCN, DropEdge with 20\% edge drop rate, and SoftEdge with 20\% soft edges. 
	} 	

%\vspace{-3mm}
\label{fig:conv}
\end{figure}
%\end{wrapfigure}

Consider, for example, the  established benchmark  NCI1 dataset~\citep{Morris+2020}, where
 each input graph represents a chemical compound with  nodes and edges respectively denoting atoms and bonds in a molecule. For such dataset, one can model the Weisfeiler-Lehman graph isomorphism test by comparing the histograms of the node colors of  graph pairs in the dataset to demonstrate the activity cliff phenomena. Figure~\ref{fig:similarity:nci1} presents the cosine similarity scores of such histograms for all graph pairs  with {\em opposite labels} both in the original graphs and in the  original graphs with randomly dropping 40\% of edges.  
This figure shows that in the original graphs about 5.85\% of the graphs with a structure similarity of over 0.9 but with opposite labels, and the percentage increases to 24.21\% if we use a cosine similarity threshold of 0.8. 
Notably, after the random edge dropping, as shown in Figure~\ref{fig:similarity:nci1}, the percentages of graphs with cosine similarity scores over 0.9 and 0.8 increase to 6.89\% and 34.08\%   respectively. 
These observations indicate that, in the NCI1 dataset there are many graphs  have very similar structures but with different labels, and node and edge modifications may change the semantics of the original graphs and flip the graph labels. 
We note that 
in the above motivation analysis, we assume that  all edges and nodes are equally important. There also exist scenarios that some graphs are not similar, but changing their key nodes/edges can dramatically change these graphs' semantic meanings as well, such as removing the 
hub nodes in social  graphs. 
These observations suggest that  graph augmentation should be carefully  conducted, and an overaggressive graph augmentation approach can easily result in under-fitting, as will be elaborated next.

\begin{figure}%[th]
%\vspace{-4mm}
	\centering
	{\includegraphics[width=0.304\textwidth]{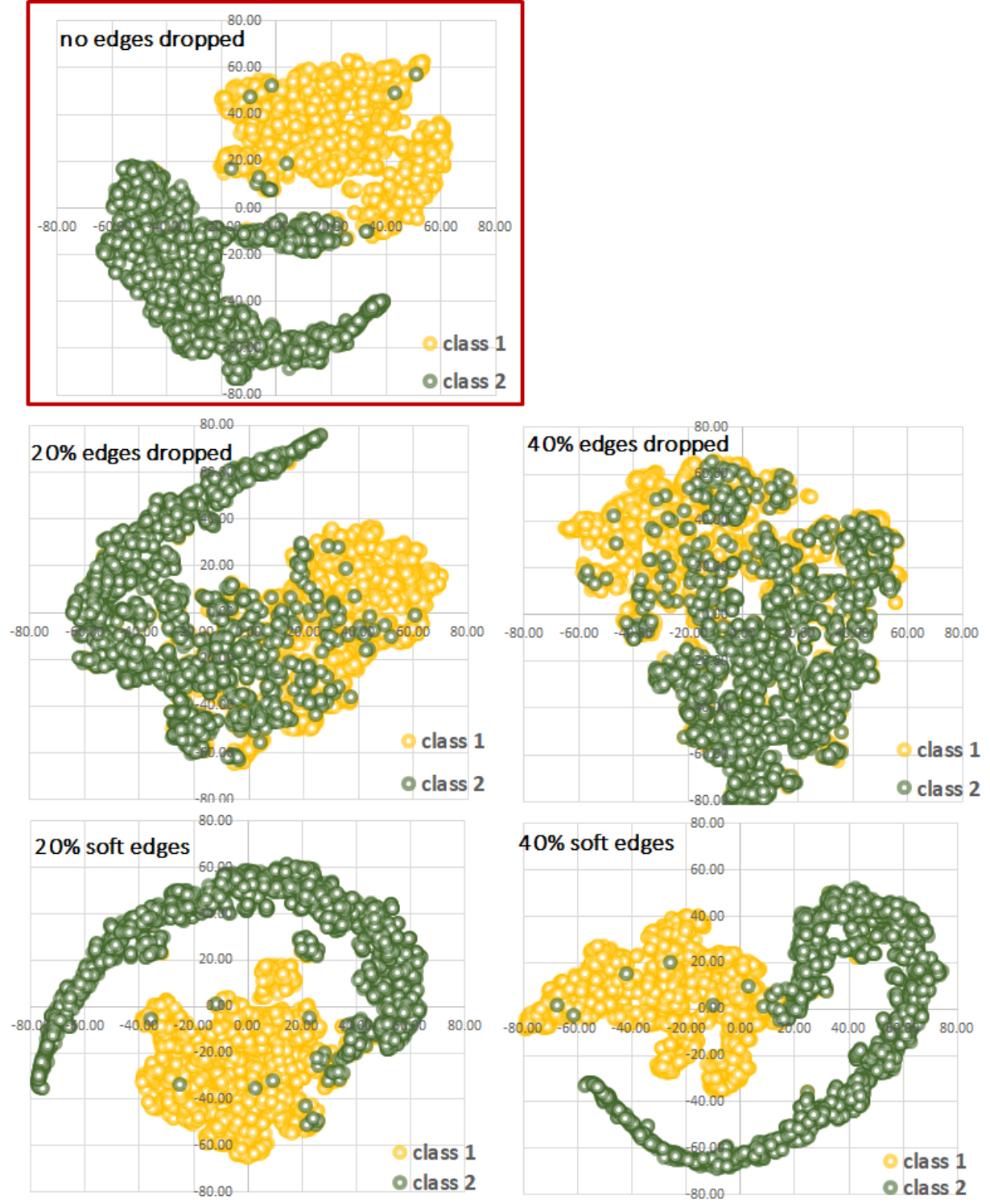}}	
%\vspace{-5mm}
	\caption{The original training graphs' 2D embeddings generated by the trained model. Top: 
	ResGCN; Middle: 
	DropEdge with  edge drop rate of  20\% and 40\%. Bottom:  
	SoftEdge with soft edge rate of  20\% and 40\%.
}	
%\vspace{-9mm}
\label{fig:dropedgeintrusion}
\end{figure}

In fact, a form of overaggressive regularization has been observed 
in the popular edge manipulation method DropEdge~\citep{rong2020dropedge}, which randomly removes a portion of the edges in a given graph for  augmentation. In specific, Figure~\ref{fig:conv} plots the training loss (top subfigure) and
validation accuracy (bottom subfigure) of the ResGCN~\citep{li2019deepgcns} and DropEdge (with 20\% edge drop rate) methods with 8 layers across the 350 training epochs on the NCI1  dataset. From Figure~\ref{fig:conv}, we can see that ResGCN over-fits the training dataset with close to zero loss, while DropEdge  under-fits the training data with a large loss, although their validation accuracy is not too bad. 
The overaggressive regularization behavior can be further confirmed by the graph embeddings formed. 
Figure~\ref{fig:dropedgeintrusion} pictures the  original training graphs' embeddings,  generated by the trained model of ResGCN as well as DropEdge with 20\% and  40\% drop rate. We project them to  2D using t-SNE~\citep{vandermaaten08a}. 
From the top, we can see that, the trained model {\em without} DropEdge can embed  the training graphs into two separable regions for the  two classes of the dataset. 
From the middle subfigures,  we can see that, the model trained {\em with} DropEdge projects the original training samples to very overlapped embedding regions of the two classes;  
as the drop rate increases the overlapped  regions enlarge. 
This indeed indicates a form of under-fitting.

%\begin{wrapfigure}{r}{0.304\textwidth}
%\end{wrapfigure}

To address  the aforementioned overaggressive  augmentation issue 
\textcolor{black}
{caused by dropping/adding graph edges and nodes 
}
we propose \soft, which simply assigns random  weights in (0, 1) to a portion of the edges of a given graph to generate synthetic graphs. 
Different to that resulting from DropEdge and DropNode, the synthetic graph generated by \soft$ $ 
has the same nodes and their connectivities as the original graph, thus mitigating the semantic changes during graph  augmentation.  As shown in Figures~\ref{fig:conv} (green curve) and ~\ref{fig:dropedgeintrusion} (the bottom subfigures), 
\textcolor{black}{
SoftEdge maintains a good trade-off between under-fitting in DropEdge and over-fitting in ResGCN, serving as an effective regularizer for graph classification.  }

  In the experiment section, we further   demonstrate, using several benchmark graph classification datasets, that  SoftEdge can effectively regularize  graph classification learning, resulting in
 superior  accuracy to popular  node and edge manipulation approaches and notable resilience to the accuracy degradation in deeper GNNs.

%\iffalse
The \soft$ $ is easy to be implemented in practice. 
For example, it can be implemented with the following  6 lines of code in PyTorch Geomertic: 
%\vspace{-2mm}
\scriptsize
\begin{lstlisting}
1. row, _ = data.edge_index #data is the input graphs
2. p = 0.2  # 20%  of edges with soft weights
3. softedge = (1- torch.rand((row.size(0),))).to(device) 
4. mask = data.edge_index.new_full(
        (row.size(0), ), 1 - p, dtype=torch.float)
5. mask = torch.bernoulli(mask)
6. data.edge_weight = softedge*(1-mask)+mask 
\end{lstlisting}
\normalsize
%\vspace{-4mm}
%\fi

%\vspace{-2mm}
\section{Related Work}
%\vspace{-2mm}
Graph data augmentation~\citep{zhao2021data,arxiv.2202.08235} has been shown to be very effective in regularizing GNNs for generalizing  to unseen graphs~\citep{kipf2017semi,YingY0RHL18,velickovic2018graph,klicpera_diffusion_2019,xu2018how,bianchi2020mincutpool}.  Nonetheless, graph data augmentation  is rather under-explored due to the  arbitrary structure and topology in graphs. Most of such strategies  heavily focus on perturbing nodes and edges in   graphs~\citep{10.5555/3294771.3294869,zhang2018bayesian,rong2020dropedge,ChenLLLZS20,3412086zhou,10.1145/3394486.3403168,WangWLCLH20,Fu2020TSExtractorLG,abs-2009-10564,abs-2104-02478,zhao2021data}. For example, DropEdge~\citep{rong2020dropedge} randomly removes a set of edges of a given graph.  %GAUG~\citep{zhao2021data} learns to perturb graph edges for node classification.
DropNode, representing node sampling based
methods~\citep{10.5555/3294771.3294869,abs-1801-10247,10.5555/3327345.3327367},  samples a set of nodes from a given graph.  DropGNN~\citep{papp2021dropgnn} executes multiple GNN runs on the
input graph, each with random node dropping. 
\textcolor{black}
{
Leveraging sample pairs under the context of Mixup~\citep{MixUp17,GuoMZ19,Guo_2020} for graph data augmetation has also been shown to be very promising, including Graph transplant~\cite{ParkSY22}, GraphMixup~\citep{mixupgraph} and ifMixup~\cite{guo2022intrusionfree}.  
}
\textcolor{black}
{
Compared to the existing works,
our study  identifies an intrinsic problem in graph data augmentation, 
%with simple node and edge manipulations, 
namely overaggressive graph augmentation causing under-fitting, which in turn inspires us to devise a novel method, without  edge or node deletion/addition, to address this  issue in graph augmentation. 
According to~\citep{Ding08235,zhao2022graph}, the data augmentaiton setting of our method falls in the following category: supervised learning, graph-level classification, edge and node removal and addition. 
}

\textcolor{black}{
Graph data augmentation has also been successfully used for contrastive learning ~\citep{chen2020simple,he2020momentum,You2020GraphCL,you2021graph,2020arXiv201014945Z,suresh2021adversarial,SunXWCZ21}. These methods aim at learning informative graph representations under a self-supervised learning setting. In contrast, we here target a supervised  learning setting, aiming to cope with the overaggressive graph augmentation  caused by dropping or adding graph edges and nodes. 
}

 Our work here is also related to the diffusion processing on the graph adjacency matrix~\citep{DBLP:journals/corr/abs-1911-05485,HassaniK2020,zhao2021adaptive,chamberlain2021grand}. 
 Diffusion  is a fundamental principle which has been applied to many domains such as image, graph, and physics etc. 
Compared to the adjacency matrix which focuses on characterizing local information, the diffusion matrix tends to capture global information and needs to be carefully designed. 
In contrast, instead of capturing the global information as in diffusion models, the equivalent adjacency matrix in SoftEdge tries to retain as much semantics of the original graph as possible by a simple design.  Also, leveraging diffusion process to change edge weight would involve careful design of a new learning paradigm, which is in contrast to the simplicity of our proposed method.

%\vspace{-1mm}
\section{Augmented Graph with Random Soft Edges}
%\vspace{-2mm}
\subsection{Graph Classification with Soft Edges}
%\vspace{-1mm}
%\textbf{Graph Classification} 
We consider an undirected graph 
 $G=(V,E)$ with the node set $V$ and the edge set $E$.  
$\mathcal{N} (v)$ represents the neighbours of node $v$, namely the set of nodes in the graph that are directly connected to $v$ in $E$ (i.e., \{$u \in V| (u,v) \in E, v \in V\}$). 
We denote with $N = |V|$ the number of nodes and $A \in\mathbf{R}^{N \times N}$ the adjacency matrix. 
Each node  in $V$ is also associated with a $d$-dimensional  feature vector, forming the feature matrix $X \in R^{|V| \times d}$ of the graph. 
Consider the graph classification task with $C$ categories. Based on a GNN, we aim to learn a mapping function $f: G 	\rightarrow y$, which assigns a graph $G$ to one of  the $C$-class labels $y \in \{1, \dots, C\}$.  

Modern GNNs leverage both the graph structure and node features to construct a distributed vector to represent a graph. The forming of such graph representation or embedding follows the ``message passing'' mechanism for the neighborhood aggregation. In a nutshell,  
every node starts with the embedding given by its initial features $X(v)$. A GNN  then iteratively updates the embedding of a node  $h_{v}$ by aggregating representations (i.e., embeddings) of its neighbours layer-by-layer,  keep transforming the node embeddings. After the construction of  individual node embeddings, the entire graph representation $h_{G}$ can then be obtained through a READOUT function, which  aggregates the embeddings of all nodes in the graph.  

Formally, the $v$-th node's representation $h_{v}^{k}$ at the $k$-th layer of a GNN is constructed as follows: 
\begin{equation}
\label{gcnAgg}
  h_{v}^{k} = \text{AGGREGATE} (h_{v}^{k-1}, {h_{u}^{k-1}|u \in \mathcal{N}(v)}, W^{k}) , 
\end{equation}
where  $W^{k}$ is the trainable weights at the $k$-th layer; \text{AGGREGATE} denotes an aggregation function implemented by the specific GNN model (e.g.,  the permutation invariant pooling operations Max, Mean, Sum); and $h_{v}^{0}$ is typically initialized as the node input feature $X(v)$.

To provide the graph level representation $h_{G}$, a GNN typically  aggregates node representations $h_{v}$  by implementing a READOUT graph pooling function (e.g., Max, Mean, Sum) to summarize information from its individual nodes:
\begin{equation}
\label{readout}
    h_{G} = \text{READOUT}({h_{v}^{k}|v \in V}). 
\end{equation}
The final step of the graph classification is to map the graph representation $h_{G}$ to a graph label $y$, through, for example, a softmax layer.

%\textbf{Graphs with Soft Edges} 
Our proposed \soft$ $ leverages graph data with  soft edges, which  
requires the GNN networks be able to take the edge weights into account for message passing when  implementing  Equation~\ref{gcnAgg} to generate node representations. Fortunately, the two  popular GNN networks, namely GCNs~\citep{kipf2017semi} and GINs~\citep{xu2018how}, can naturally take into account the soft edges through weighted summation of neighbor nodes, as follows.

In a GCN, the  adjacency matrix can naturally include edge weight values between zero and one~\citep{kipf2017semi}, instead of binary edge weights. Consequently, Equation~\ref{gcnAgg}  in a GCN is implemented through a weighted sum operation: 
\begin{equation}
\label{gcnagg}
\mathbf{h}^{k}_v = \sigma \left( W^k \cdot \left( \sum_{u \in \mathcal{N}(v) \cup
\{ v \}} \frac{e_{u,v}}{\sqrt{\hat{d}_u \hat{d}_v}} \mathbf{h}^{k-1}_{u} \right) \right),
\end{equation}
where  $\hat{d}_v = 1 + \sum_{u \in \mathcal{N}(v)} e_{u,v}$;  $e_{u,v}$ is the edge weight between nodes $u$ and $v$; $W^{k}$ stands for the trainable weights at layer  $k$; and $\sigma$ is the non-linearity transformation ReLu.

Similarly,  to handle soft edge weights in a GIN, we can simply replace the sum operation of the isomorphism operator with a weighted sum calculation, and get the following implementation of Equation~\ref{gcnAgg} for message passing: 
\begin{equation}
\label{ginagg}
\mathbf{h}^{k}_v = \text{MLP}^{k} \left( (1 + \epsilon^{k}) \cdot
\mathbf{h}_v^{k-1} + \sum_{u \in \mathcal{N}(v)} e_{u,v} \cdot \mathbf{h}_u^{k-1} \right). 
\end{equation}
Here,  $\epsilon^{k}$ denotes a learnable parameter.

\subsection{Graph Learning with Random Soft Edges}
\label{mixschema}

Our method \soft$ $ takes as input a graph
$G=(V,E)$, and a given  $ \lambda \in (0, 1)$ representing the percentage of edges that will be associated with random soft weights.
Specifically,  at each training epoch, \soft$ $ randomly selects  $ \lambda$ percentage  of edges in $E$, and assigns each of the selected edges (denoted as $\widetilde{E}$, $\widetilde{E} \subset E$) with a random weight $w_{\lambda}$ that is uniformly sampled from $(0, 1)$.

Assume that all edge weights of the original graph are  $1$.
With the above operation the newly created synthetic graph 
will have the same nodes (i.e., $V$) and the edge set (i.e., $E$)   as the original graph $G$, except that the edge weights are not all with the value $1$ anymore. 
In specific,  $\lambda$\% of edges will be associated with a random 
weight in $(0, 1)$, and the rest  will be unchanged. Formally, the edge weights in \soft $ $ are as follows: 
%\vspace{-2mm}
\begin{equation}
\label{softedge}
e(u, v)_{w_{\lambda}} = 
\left\{
\begin{array}{cl}
     w_{\lambda}, & {\rm if} (u, v) \in \widetilde{E}  \\
     1 & {\rm otherwise}   
\end{array}
\right.,
\end{equation}
%%\vspace{-2mm}
where $w_{\lambda} \in (0, 1)$, $u, v \in V$, and $\widetilde{E} \subset E$.

\begin{figure}[h]
%\vspace{-3mm}
	\centering	
	{\includegraphics[width=0.48\textwidth]{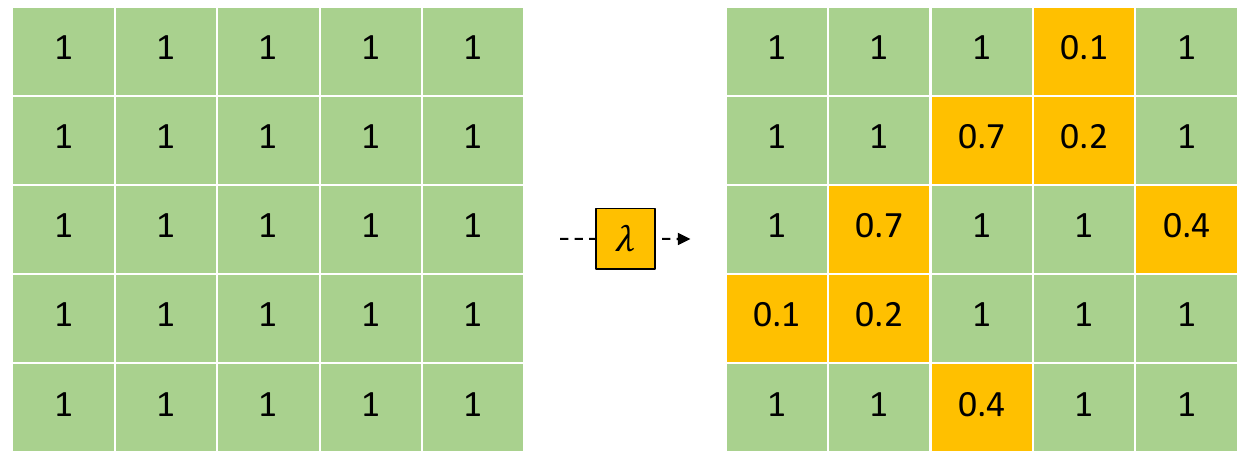}}	
%\vspace{-2mm}
	\caption{Illustration of  \soft$ $: The left is the original adjacency matrix (self-loop; fully connected) with binary weights, and the right is the \soft's adjacency matrix where $\lambda$\% of its elements  are in  $(0, 1)$.}
\label{fig:schema}
\end{figure}

This softening edges process is illustrated in Figure~\ref{fig:schema}, where the left subfigure is the original adjacency matrix $A$ for a fully connected graph with self-loop, and the right is the new adjacency matrix $A_{w_{\lambda}}$ generated by \soft, which can be obtained by the dot product between $A$ and the edge weights.
%$E_{w_{\lambda}}$.  
The yellow cells are edges with  soft weights (belong to edges in $\widetilde{E}$).  
With such weighted edges, message passing in GNNs as formulated in Equations~\ref{gcnagg} and ~\ref{ginagg} can be directly applied for learning. 

The pseudo-code of the synthetic graph generation in \soft$ $  is described in Algorithm~\ref{alg:softedge}.
\small
\begin{algorithm}%[h]
   \caption{Synthetic graph generation in \soft}
   \label{alg:softedge}
\begin{algorithmic}
   \State  {\bfseries Input:} a graph  $G=(V,E)$ (all edges in  $E$ have weight 1); percentage of soft weights $\lambda \in (0, 1)$ 
\State  {\bfseries Output:} a synthetic graph $G_{w_{\lambda}}$
   \State  Sample $\lambda$ percentage of edges from $E$, forming $\widetilde{E}$
\For {$(u, v) \in  E$}
\If{$ (u, v) \in \widetilde{E}$}
   \State  Sample $w_{\lambda}$ from $(0, 1)$ uniformly
\State  $e(u, v)_{w_{\lambda}}=w_{\lambda}$
\Else
\State  $e(u, v)_{w_{\lambda}}=1$
\EndIf
 \EndFor  
\State  Return synthetic graph with new edge weights $e(u, v)_{w_{\lambda}}$
\end{algorithmic}
\end{algorithm}
\normalsize

%\begin{wrapfigure}{r}{0.5\textwidth}
%\vspace{-2mm}
%\end{wrapfigure}

%\vspace{-4mm}
\subsection{Discussion}
%\vspace{-2mm}

\textbf{Relation to DropEdge~\citep{rong2020dropedge}} 
The proposed \soft$ $ method is inspired by and  closely related to DropEdge. Unlike DropEdge, \soft$ $ excludes the edge deletion, and instead  assigns random weights between $(0, 1)$ to a portion of the edges in a graph (\soft$ $ would be equivalent to DropEdge if the soft weights were all zero).
As a consequence, different from DropEdge,  \soft$ $ 
 maintains the same nodes and their connectivities as the original graphs in its augmented graphs, thus effectively mitigating the semantic changes during the graph augmentation process. 

\textbf{Why \soft$ $ works?} 
Intuitively,  GNNs essentially update node embeddings with  a weighted sum of the neighborhood information through the graph structure. In \soft$ $, edges of the graph are associated with different weights that are randomly sampled. Consequently, 
 \soft$ $ enables a random subset aggregation instead of the fixed neighborhood aggregation during the learning of GNNs,  which
provides dynamic  neighborhood over the graph for message passing. 
This can be considered as a form of data augmentation or denoising filter, which in turn helps  the graph learning because edges in real graphs are often noisy and arbitrarily defined. 

More importantly,  different from DropEdge, the synthetic graph generated by \soft$ $ 
has the same nodes and their connectivities (i.e., the same $V$ and $E$) as the original graph. The only difference  is that the synthetic graph has some soft edges. 
As such, the synthetic graphs maintain a large similarity to
 their corresponding original  graphs and alleviate the semantic changes during the graph augmentation, which we believe   attributes to the superiority  of  \soft $ $ to DropEdge and DropNode.

\textcolor{black}{
\textbf{\soft$ $ Improves GNN Expressiveness} 
%\soft$ $ improves the expressiveness of GNN.
\begin{lem}
\label{lem3}
%\vspace{-3mm}
When two different graphs are indistinguishable by the Weisfeiler-Lehman test,  assigning random weights to the graphs resolves the ambiguities leading to the failure of the Weisfeiler-Lehman test. 
\end{lem}
%\vspace{-2mm}
\noindent {\em Proof:}
Due to the random soft edges, each message in the two graphs is transformed differently depending on the weight between the endpoint nodes. 
As such, the GNN produces the same embedding for the two graphs if we have exactly the same subset of edges with the same random weights, which has the probability  zero due to the following fact in \soft:  each $w_{\lambda}$  is independently drawn from a continuous distribution over $(0, 1)$; hence, the probability of  two sets of soft weights $ \widetilde{E}_{w_{\lambda}^{i} }$ and $\widetilde{E}_{w_{\lambda}^{j} }$ being identical  is  zero. 
Intuitively,  the exploiting of graph structure in message passing GNNs can be understood as the breadth first search tree~\citep{xu2018how}. Such trees will not be identical when associating  some random weights in different tree traversal paths.  
\hfill $\Box$
}

\textcolor{black}{
\textbf{Sampling in Uniform Distribution} It is worth noting that, there is no need to force the soft edge weights to be sampled from a Uniform distribution. For example, one can use a Beta distribution or a Truncated Gaussian. 
}

%%\vspace{-1mm}
%\begin{wrapfigure}{r}{0.4\textwidth}
\begin{figure}[h]
%\vspace{-5mm}
	\centering
	{\includegraphics[width=0.4\textwidth]{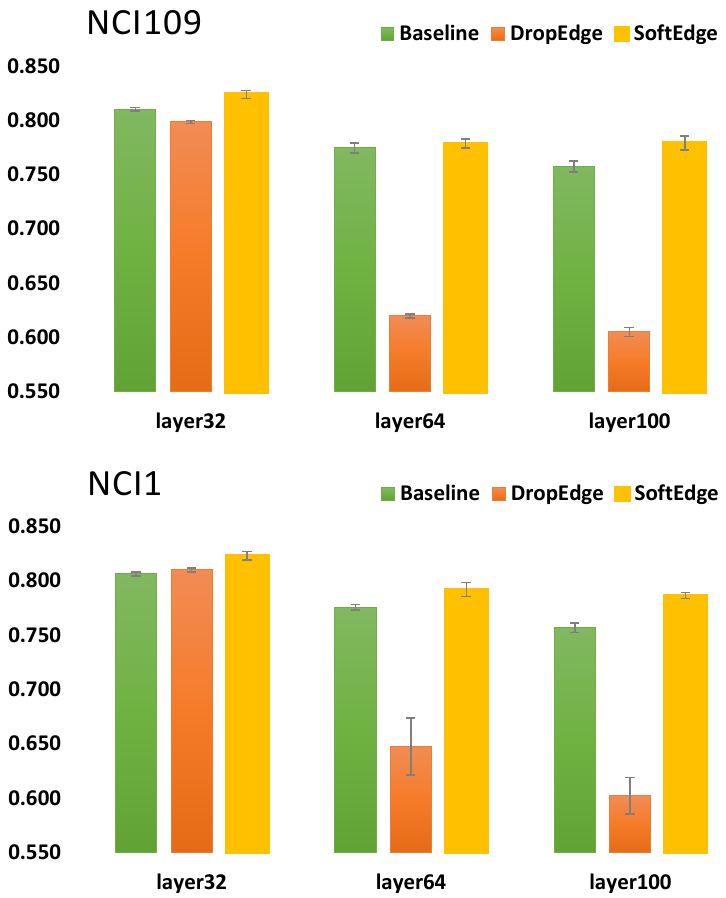}}	
%%\vspace{-7mm}
	\caption{Accuracy obtained by ResGCN, DropEdge, and \soft$ $  with  32, 64, and 100 layers on  NCI109 and NCI1. }
	\label{fig:dropedgeDeeper}
%%\vspace{-14mm}
\end{figure}
%\end{wrapfigure}
%%\vspace{-1mm}

%%\vspace{-2mm}

%\vspace{-3mm}
\section{Experiments}

%\vspace{-2mm}

\subsection{Settings}
\label{setting}
 %\vspace{-2mm}

\textbf{Datasets}
We conduct experiments 
using six  graph classification tasks from the  graph benchmark datasets  collection TUDatasets~\citep{Morris+2020}:  PTC\_MR, NCI109, NCI1, and MUTAG for small molecule classification, and ENZYMES and PROTEINS for protein categorization.  
 These  datasets have been widely used for benchmarking such as in~\cite{xu2018how} and 
can be downloaded directly using PyTorch Geometric~\citep{Fey/Lenssen/2019}'s built-in function online~\footnote{https://chrsmrrs.github.io/datasets/docs/datasets}. 
Table~\ref{tab:data} %in the Appendix
summarizes the  statistics of the datasets, including the number of graphs, the average node number per graph,  the average edge number per graph, the number of node features, and the number of classes. 

%\iffalse
\begin{table}[h]%{r}{0.66\textwidth}
  \centering
\scalebox{0.9}{
\begin{tabular}{l|l|c|c|c|c}\hline
Name &	graphs&	nodes &	edges& features& classes	\\ \hline

PTC\_MR&	334	&14.3&	29.4&	18&	2		\\
NCI109&	4127&	29.7&	64.3&	38&	2		\\
NCI1&	4110&	29.9&	64.6&	37&	2		\\
MUTAG	&188&	17.9&	39.6&	7&	2		\\
ENZYMES	&600	&32.6&	124.3&	3&	6		\\
PROTEINS&	1113&	39.1&	145.6&	3&	2 \\
\hline
\end{tabular}
}
%%\vspace{-3mm}
\caption{Statistics of the graph classification benchmark datasets. 
  }   
  %	 %\vspace{-5mm}
  \label{tab:data} 
%\end{table} 
\end{table}
%\vspace{-3mm}
%\fi

\textbf{Comparison Baselines} 
We compare our method with three baselines:  DropEdge~\citep{rong2020dropedge}, DropNode~\citep{10.5555/3294771.3294869,abs-1801-10247,10.5555/3327345.3327367},  and Baseline. 
For the Baseline model, we use two popular GNN network architectures: GCNs~\citep{kipf2017semi} and GINs~\citep{xu2018how}. 
GCNs use spectral-based convolutional operation to learn spectral features of graph through message aggregation, leveraging a normalized adjacency matrix. 
In the experiments, we use the GCN with  Skip Connection~\citep{7780459} as that in~\citep{li2019deepgcns}.  Such Skip Connection empowers the GCN to benefit from deeper layers in a GNN. We denote this GCN as ResGCN. 
The GIN  represents the state-of-the-art GNN  architecture. It leverages the nodes' spatial relations to aggregate neighbor  features. 
For both ResGCN and GIN, we use their implementations in the PyTorch Geometric platform~\footnote{https://github.com/pyg-team/pytorch\_geometric}.

We note that, in this paper, we aim to 
study the overaggressive graph augmentation issue in edge and node manipulation. 
Therefore, we compare our method with commonly used data augmentation baselines DropEdge and DropNode. We believe that, our method is also useful for advanced graph data augmentation strategies, which we will leave for future studies.

\textbf{Detailed Settings} 
We follow the evaluation protocol and hyperparameters search of GIN~\citep{xu2018how} and DropEdge~\citep{rong2020dropedge}. In detail,  we evaluate the models  using 10-fold cross validation, and compute the mean and standard deviation of three  runs. 
 Each fold is trained with 350 epochs with AdamW optimizer~\citep{KingmaB14}. The initial learning rate is decreased by half every 50 epochs. 
 The hyper-parameters searched for  all models on each dataset are as follows:  (1) initial learning rate $\in$ \{0.01, 0.0005\}; 
 (2) hidden unit of size  64; (3)  batch size $\in$ \{32, 128\}; (4)  dropout ratio  after the dense layer $\in$ \{0, 0.5\};  (5) drop ratio in DropNode and DropEdge  $\in$ \{20\%, 40\%\},   (6) number of layers in GNNs $\in$ \{3, 5, 8, 16, 32, 64, 100\}.
For \soft, $\lambda$, the percentage of soft edges, is 20\%, and the soft edge weights are uniformly sampled from (0, 1), unless otherwise specified. 
 Following GIN~\citep{xu2018how} and DropEdge~\citep{rong2020dropedge}, we  report the case giving
 the best 10-fold average cross-validation accuracy. 
Our  experiments use a NVIDIA V100/32GB GPU.

%	 %\vspace{-2mm}
%\subsection{Main Results}

\begin{table*}[h]
  \centering
   \scalebox{0.95}{
\begin{tabular}{ll||c|c|c|c||c}\hline
Dataset&Method&3 layers&			5 layers 	&	8 layers 	&	16 layers		&max	 \\ \hline
PTC\_MR&ResGCN	&0.619$\pm$0.006&0.642$\pm$0.003&0.638$\pm$0.003&0.652$\pm$0.008&0.652 \\
& DropEdge	&0.633$\pm$0.006&0.653$\pm$0.007&0.646$\pm$0.002&0.652$\pm$0.005&0.653 \\
& DropNode	&0.620$\pm$0.002&0.648$\pm$0.018&0.642$\pm$0.005&0.649$\pm$0.007&0.649 \\

&\soft	&\textbf{0.649$\pm$0.006}&\textbf{0.665$\pm$0.004}&\textbf{0.664$\pm$0.006}&\textbf{0.671$\pm$0.006}&\textbf{0.671}\\\hline
NCI109&ResGCN	&\textbf{0.791$\pm$0.004}&0.803$\pm$0.003&0.807$\pm$0.001&0.810$\pm$0.003&0.810 \\
& DropEdge	&0.760$\pm$0.000&0.778$\pm$0.008&0.800$\pm$0.002&0.808$\pm$0.003&0.808 \\
& DropNode	&0.765$\pm$0.002&0.793$\pm$0.015&0.801$\pm$0.002&0.802$\pm$0.001&0.802 \\

&\soft	&0.790$\pm$0.005&\textbf{0.813$\pm$0.003}&\textbf{0.821$\pm$0.001}&\textbf{0.824$\pm$0.001}&\textbf{0.824}\\\hline
NCI1&ResGCN	&0.796$\pm$0.002&0.804$\pm$0.003&0.810$\pm$0.002&0.814$\pm$0.003&0.814 \\
& DropEdge	&0.776$\pm$0.001&0.795$\pm$0.010&0.814$\pm$0.001&0.818$\pm$0.001&0.818 \\
& DropNode	&0.778$\pm$0.001&0.805$\pm$0.019&0.812$\pm$0.001&0.813$\pm$0.002&0.813 \\

&\soft	&\textbf{0.799$\pm$0.002}&\textbf{0.819$\pm$0.001}&\textbf{0.822$\pm$0.002}&\textbf{0.827$\pm$0.001}&\textbf{0.827}\\\hline
MUTAG&ResGCN	&0.827$\pm$0.003&\textbf{0.841$\pm$0.009}&\textbf{0.846$\pm$0.001}&0.846$\pm$0.000&0.846 \\
& DropEdge	&0.816$\pm$0.003&0.832$\pm$0.003&0.850$\pm$0.006&0.858$\pm$0.003&0.858 \\
& DropNode	&0.823$\pm$0.006&0.829$\pm$0.006&0.839$\pm$0.003&0.858$\pm$0.006&0.858 \\

&\soft	&\textbf{0.859$\pm$0.008}&\textbf{0.841$\pm$0.001}&\textbf{0.846$\pm$0.005}&\textbf{0.874$\pm$0.003}&\textbf{0.874}\\\hline
ENZYMES&ResGCN	&0.508$\pm$0.015&0.537$\pm$0.003&0.540$\pm$0.010&0.541$\pm$0.014&0.541 \\
& DropEdge	&0.489$\pm$0.003&0.521$\pm$0.003&0.564$\pm$0.011&0.597$\pm$0.002&0.597 \\
& DropNode	&0.510$\pm$0.002&0.532$\pm$0.006&0.573$\pm$0.006&0.590$\pm$0.004&0.590 \\

&\soft	&\textbf{0.518$\pm$0.004}&\textbf{0.564$\pm$0.005}&\textbf{0.586$\pm$0.005}&\textbf{0.615$\pm$0.004}&\textbf{0.615}\\\hline
PROTEINS&ResGCN	&0.738$\pm$0.005&0.738$\pm$0.002&0.739$\pm$0.003&0.747$\pm$0.005&0.747 \\
& DropEdge	&\textbf{0.747$\pm$0.003}&\textbf{0.750$\pm$0.003}&\textbf{0.749$\pm$0.002}&0.755$\pm$0.002&0.755 \\
& DropNode	&0.745$\pm$0.004&0.748$\pm$0.001&0.744$\pm$0.002&0.745$\pm$0.005&0.748 \\

&\soft	&0.745$\pm$0.002&0.748$\pm$0.000&0.741$\pm$0.000&\textbf{0.757$\pm$0.003}&\textbf{0.757}\\
\hline
\end{tabular}
}
  	% %\vspace{-1mm}
\caption{Accuracy  of the testing methods with ResGCN networks as baseline.  We report mean accuracy over 3 runs of 10-fold cross validation with standard deviations (denoted $\pm$). 
Max depicts the max accuracy over different GNN layers. 
 Best results are  in \textbf{Bold}. 
  }  
  %	 %\vspace{-3mm}
  \label{tab:accuracy:gcn} 
\end{table*}

%%\vspace{-3mm}
\subsection{Results with  ResGCN}
%	 %\vspace{-1mm}
\subsubsection{Main Results}
%%\vspace{-1mm}
Table~\ref{tab:accuracy:gcn} presents the accuracy obtained by the ResGCN~\citep{kipf2017semi,li2019deepgcns} baseline,  DropEdge, DropNode, and \soft$ $   on the six  datasets, where we evaluate GNNs with 3, 5, 8, and 16 layers. In the table,  the best results  are in \textbf{Bold}.

Results in the last column of Table~\ref{tab:accuracy:gcn} show that \soft$ $ outperformed all the three comparison models on all the six datasets when considering the max accuracy obtained with layers 3, 5, 8, and 16. 
For example, when compared with  ResGCN, \soft$ $ increased the accuracy from 65.2\%,  84.6\%, and 54.1\%  to 67.1\%, 87.4\%, and 61.5\%, respectively, on the PTC\_MR, MUTAG, and ENZYMES datasets. Similarly, when compared with DropEdge, \soft$ $ improved the accuracy from 65.3\%, 80.8\%, and 84.6\% to 67.1\%, 82.4\%, and 87.4\%, respectively, on the PTC\_MR, NCI109, and MUTAG datasets.

Promisingly, as highlighted in bold  in the table, \soft$ $  obtained superior accuracy to the other comparison models in most settings regardless of the network layers used. For example, on the PTC\_MR, NCI1, and ENZYMES datasets, \soft$ $ outperformed all the  three baselines with all the  network depths tested (i.e., 3, 5, 8, and 16 layers).

Another  observation here is that  \soft$ $  improved or  maintained  the predictive accuracy in most of the cases as the networks increased their depth  from 3, 5, 8, to 16.  As can be seen in Table~\ref{tab:accuracy:gcn}, the best accuracy of \soft$ $ on all the six datasets were obtained by \soft$ $ with  16 layers.

\subsubsection{Ablation Studies}
\label{ablactionstu}
%%\vspace{-2mm}

We conduct  ablation studies to evaluate  \soft$ $. We  particularly compare our strategy with DropEdge, since it is the most related algorithm to our approach. 

\textbf{Effect on GNN Depth} 
We  conduct experiments on further increasing the networks' depth by adding more layers, including ResGCN with 32, 64, and 100 layers. 
Results on the NCI109 and NCI1 datasets are presented in Figure~\ref{fig:dropedgeDeeper}.

Results in Figure~\ref{fig:dropedgeDeeper} show that DropEdge  significantly degraded the predictive accuracy when ResGCN has 64 and 100 layers; and surprisingly, the  baseline model ResGCN  performed better than  DropEdge as the networks went deeper. Notably, the \soft$ $ method was able to slow down the degradation in terms of the accuracy obtained with deeper networks, outperforming both baselines ResGCN and  DropEdge  with layers 32, 64, and 100.

%\begin{wrapfigure}{r}{0.4\textwidth}
\begin{figure}[h]
%%\vspace{-3mm}
	\centering
	{\includegraphics[width=0.4\textwidth]{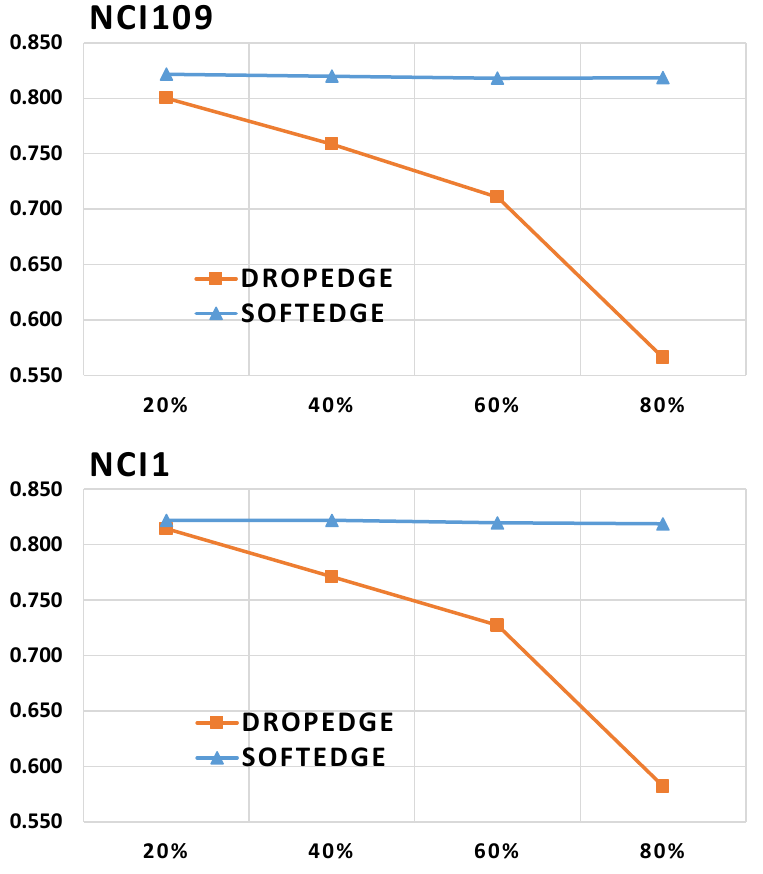}}	

%%\vspace{-2mm}
	\caption{Accuracy obtained by DropEdge and \soft$ $ as ratio of modified edges  (x-axis) increasing on  NCI109 and NCI1. }	

%%\vspace{-4mm}
\label{fig:nci1droprate}
\end{figure}
%\end{wrapfigure}

\textbf{Ratio of Modified Edges}  
In Figure~\ref{fig:nci1droprate}, we present the results when varying the percentage of modified edges (i.e., $\lambda$) in both DropEdge and \soft$ $ with 8 layers. We evaluate the percentage of 20\%, 40\%, 60\%, and 80\%, respectively, on  NCI109 and NCI1. We chose these two datasets since they have the largest number of  node features amongst the six tested datasets, which are more challenging. This is because, intuitively with a large number of node features the overaggressive augmentation % \collision$ $
issue could be mitigated. 

Results in Figure~\ref{fig:nci1droprate} show that,
DropEdge was very sensitive to the percentage of modified edges as the model's predictive accuracy decreases significantly with the drop rate.
This is expected, as DropEdge completely dropped selected edges the graph structure would be largely changed especially when the drop rate is high. 
On the other hand, \soft$ $ seemed to be robust to the percentage of edges being modified as the accuracy obtained for the candidate ratios 
was quite  similar. This is mainly because, unlike DropEdge, the synthetic graphs generated by \soft$ $ remain a large similarity to the original graphs as the node structures are kept the same.

%%\vspace{-2mm}
\begin{table*}[h]
  \centering
   \scalebox{1}{
\begin{tabular}{ll|c|c|c|c}\hline
Dataset&Method&			5 layers 	&	8 layers 	&	16 layers		&max	 \\ \hline
PTC\_MR&GIN&0.647$\pm$0.005&0.659$\pm$0.003&0.673$\pm$0.022&0.673 \\
& DropEdge	&0.680$\pm$0.003&0.681$\pm$0.009&0.679$\pm$0.007&0.681 \\
& DropNode	&\textbf{0.688$\pm$0.006}&0.689$\pm$0.003&0.680$\pm$0.006&0.689 \\
&\soft	&0.687$\pm$0.011&\textbf{0.691$\pm$0.009}&\textbf{0.696$\pm$0.007}&\textbf{0.696} \\ \hline

NCI109&GIN&0.818$\pm$0.002&0.823$\pm$0.002&0.820$\pm$0.001&0.823 \\
& DropEdge	&0.813$\pm$0.005&0.814$\pm$0.003&0.805$\pm$0.003&0.814 \\
& DropNode	&0.819$\pm$0.002&0.821$\pm$0.002&0.816$\pm$0.004&0.821 \\
&\soft	&\textbf{0.835$\pm$0.002}&\textbf{0.836$\pm$0.001}&\textbf{0.838$\pm$0.003}&\textbf{0.838} \\\hline

NCI1&GIN&0.820$\pm$0.002&0.821$\pm$0.001&0.821$\pm$0.002&0.821 \\
& DropEdge	&0.819$\pm$0.004&0.823$\pm$0.001&0.818$\pm$0.003&0.823 \\
& DropNode	&0.821$\pm$0.003&0.821$\pm$0.003&0.824$\pm$0.000&0.824 \\
&\soft	&\textbf{0.839$\pm$0.001}&\textbf{0.839$\pm$0.002}&\textbf{0.837$\pm$0.005}&\textbf{0.839} \\ \hline

MUTAG&GIN&\textbf{0.876$\pm$0.003}&0.871$\pm$0.003&0.874$\pm$0.000&0.876 \\
& DropEdge	&0.857$\pm$0.005&\textbf{0.878$\pm$0.009}&0.871$\pm$0.003&0.878 \\
& DropNode	&0.864$\pm$0.006&0.875$\pm$0.003&0.869$\pm$0.009&0.875 \\
&\soft	&\textbf{0.876$\pm$0.006}&\textbf{0.878$\pm$0.005}&\textbf{0.889$\pm$0.005}&\textbf{0.889} \\ \hline

ENZYMES&GIN&0.531$\pm$0.004&0.533$\pm$0.006&0.532$\pm$0.001&0.533 \\
& DropEdge	&0.480$\pm$0.003&0.497$\pm$0.013&0.487$\pm$0.004&0.497 \\
& DropNode	&0.521$\pm$0.004&0.563$\pm$0.005&0.552$\pm$0.004&0.563 \\
&\soft	&\textbf{0.571$\pm$0.002}&\textbf{0.590$\pm$0.003}&\textbf{0.588$\pm$0.009}&\textbf{0.590} \\ \hline

PROTEINS&GIN&0.741$\pm$0.006&0.744$\pm$0.005&0.743$\pm$0.001&0.744 \\
& DropEdge	&\textbf{0.742$\pm$0.002}&0.738$\pm$0.004&0.736$\pm$0.003&0.742 \\
& DropNode	&0.736$\pm$0.002&\textbf{0.745$\pm$0.002}&\textbf{0.747$\pm$0.002}&\textbf{0.747} \\
&\soft	&0.740$\pm$0.002&0.742$\pm$0.004&\textbf{0.747$\pm$0.002}&\textbf{0.747} \\ \hline

\end{tabular}
}

%\vspace{2mm}
\caption{Accuracy  of the testing methods with GIN networks as baseline.  We report mean accuracy over 3 runs of 10-fold cross validation with standard deviations (denoted $\pm$). 
Max depicts the max accuracy over different GNN layers. 
 Best results are  in \textbf{Bold}.}  

%\vspace{-6mm}
  \label{tab:accuracy:ginfull} 
\end{table*} 
%%\vspace{-2mm}

%%\vspace{3mm}
\subsection{Results with  GIN}
%%\vspace{-2mm}
In this section, we also evaluate our method using the GIN~\citep{xu2018how}  network architecture. 
Table~\ref{tab:accuracy:ginfull} presents the accuracy obtained by the GIN,  DropEdge, DropNode, and \soft$ $   on the six  datasets, where we examine  GNNs with 5, 8, and 16 layers and the best results  are in \textbf{Bold}.

Results in Table~\ref{tab:accuracy:ginfull} show that, similar to the ResGCN case,  \soft$ $ with GIN as the baseline outperformed all the other comparison models on all the six datasets, as highlighted by the last column of the table. 
For example, when compared with the GIN baseline, \soft$ $ increased the accuracy from 67.3\%,  82.1\%, and 53.3\%  to 69.6\%, 83.9\%, and 59.0\%, respectively, on the PTC\_MR, NCI1, and ENZYMES datasets. 
Similarly, when compared with DropEdge, \soft$ $ improved the accuracy from 81.4\%, 82.3\%, and 49.7\% to 83.8\%, 83.9\%,  and 59.0\%, respectively, on the  NCI109, NCI1, and ENZYMES datasets.

\textcolor{black}{
We note that 
our work here aims at studying and addressing the overaggressive graph augmentation issues caused by the widely used edge and node dropping on supervised graph classification. We conjecture that on the node level classification, we may also have graph nodes with very similar neighbourhoods but of different node labels, owing to the node and edge operation in the augmentation processes. 
Nevertheless, the aggressive augmentation issue is more complex  on node level due to the    over-smoothing phenomena~\citep{LiHW18,abs-1901-00596}, where 
the representations of all nodes in a graph may converge to a subspace that makes their  representations \textbf{ unrelated to the graph input}~\citep{LiHW18,abs-1901-00596}. 
We intend to  
study such node  classification in the future. 
}

%%\vspace{-2mm}
%\section{Discussions}
%%\vspace{-1mm}

%\vspace{-2mm}
\section{Conclusions and Future Work}
%\vspace{-2mm}
We discussed  
the overaggressive graph augmentation issue 
in simple edge  and node manipulations for graph regularization,  where the semantics of augmented graphs may differ from their original graphs, which will induce a form of under-fitting for model regularization. 
We proposed \soft, which  assigns random  weights to a portion of the edges of a given graph, enabling training GNNs with dynamic neighborhoods.  
 The synthetic graph generated by  \soft$ $ maintains the same nodes and their connectivities as the original graph, thus  mitigating the semantics changes of the original graph.  
 We also  showed that  this simple approach  resulted in  superior  accuracy to popular node and edge manipulation methods and notable resilience to the accuracy degradation with the GNN depth.

We  note that 
\soft$ $ assumes that the given graphs for training have binary edges, which is a commonly adopted  setting in the graph learning field. 
When using  graphs with real-valued edge weights,  
there are two main strategies to handle such situation in the literature, which can be adopted by our \soft$ $ algorithm. In the first method, edge weights are added to the features of the nodes~\citep{abs-2005-00687,DBLP:journals/corr/abs-2006-07739}. In the second approach, edge weights are treated the same way as node features in the Aggregation function of the GNNs~\citep{DBLP:journals/corr/GilmerSRVD17,xu2018how,kipf2017semi,hu2020strategies}. 
\textcolor{black}
{
It is also worth noting  that these two approaches can also be used by \soft$ $ for graphs that are associated with edge attributes. 
}

\bibliography{graphMixup}
\bibliographystyle{icml2022}

\end{document}